\documentclass{article}

\usepackage{amsmath}
\usepackage{graphicx,fancybox}

\usepackage[final]{Styles/neurips_2024}

\usepackage[utf8]{inputenc} 
\usepackage[T1]{fontenc}    
\usepackage{hyperref}       
\usepackage{url}            
\usepackage{booktabs}       
\usepackage{amsfonts}       
\usepackage{nicefrac}       
\usepackage{microtype}      
\usepackage{xcolor}         
\usepackage{algorithm,algorithmic}

\def \Y {\mathcal{Y}}

\def \w {\mathbf{w}}
\def \v {\mathbf{v}}

\def \x {\mathbf{x}}

\def \x {\mathbf{x}}
\def \p {\mathbf{p}}

\def \1 {\mathbf{1}}
\def \L {\mathcal{L}}

\def \u {\mathbf{u}}

\def \B {\mathcalB}

\def \x {\mathbf{x}}

\def \D {\mathcal{D}}

\def \w {\mathbf{w}}

\def \v {\mathbf{v}}
\def \M {\mathcal{M}}

\def \p {\mathbf{p}}

\def \B {\mathcal{B}}

\def \T {\mathcal{T}}

\title{Memory-Efficient Continual Learning with CLIP Models}

\author{%
  Ryan C.~King \\
  Department of Computer Science\\
  Texas A\&M University\\
  College Station, TX 77843 \\
  \texttt{kingrc15@tamu.edu} \\
  \And
  Gang Li \\
  Department of Computer Science\\
  Texas A\&M University\\
  College Station, TX 77843 \\
  \texttt{gang-li@tamu.edu} \\
  \AND
  Bobak J. Mortazavi \\
  Texas A\&M University\\
  College Station, TX 77843 \\
  \texttt{bobakm@tamu.edu} \\
  \And
  Tianbao Yang \\
  Texas A\&M University\\
  College Station, TX 77843 \\
  \texttt{tianbao-yang@tamu.edu} \\
}

\begin{document}

\maketitle

\begin{abstract}

Contrastive Language-Image Pretraining (CLIP) models excel at understanding image-text relationships but struggle with adapting to new data without forgetting prior knowledge. To address this, models are typically fine-tuned using both new task data and a memory buffer of past tasks. However, CLIP's contrastive loss suffers when the memory buffer is small, leading to performance degradation on previous tasks. We propose a memory-efficient, distributionally robust method that dynamically reweights losses per class during training. Our approach, tested on class incremental settings (CIFAR-100, ImageNet1K) and a domain incremental setting (DomainNet) adapts CLIP models quickly while minimizing catastrophic forgetting, even with minimal memory usage.
\end{abstract}

\section{Introduction}

In dynamic environments, machine learning systems must continuously learn and adapt to new information. Continual learning (CL) allows models to acquire new skills while retaining knowledge from past tasks, which is essential as data evolves over time. While there is extensive research on addressing the challenge of catastrophic forgetting in traditional supervised models, most methods—such as parameter regularization, knowledge distillation, and dynamic architectures—have not been applied to models like CLIP, which excel at understanding image-text relationships.

CLIP models need CL to adapt to real-world data streams. However, CL with CLIP models is still under-explored. Recent works, such as those by \cite{thengane2022clip} and \cite{garg2024ticclip}, have shown promising results in mitigating forgetting through rehearsal-based approaches and memory buffers. Despite these advances, a key question remains: how can we efficiently leverage memory buffers in CLIP's CL to balance new and old task performance?

Our study addresses this by proposing two approaches: one treats old and new data equally during fine-tuning, while the other dynamically reweights class losses using Distributionally Robust Optimization (DRO). We evaluate these methods in class-incremental and domain-incremental settings, demonstrating improved retention of past knowledge and efficient adaptation to new tasks with minimal memory requirements.

\section{Related Works}
\textbf{Continual Learning} There are many approaches to address catastrophic forgetting. One approach is through replay methods, which update models with a combination of new task data and examples from previous tasks stored in a memory buffer. While effective, maintaining these buffers increases computational costs and poses challenges under privacy constraints. Generative replay methods attempt to mitigate this by synthesizing prior task data, though their success depends on the quality of the generated examples. Dynamic model expansion is another technique, where architectures are extended after each task. For example, \cite{yan2021dynamically} trains a new model per task, which avoids forgetting but is impractical for large models. \cite{wang2022foster} reduces memory usage by retaining the previous model for distillation, while \cite{zhou2022model} only expands specific network blocks. Knowledge distillation (KD) is another approach, transferring knowledge from previous tasks to a target model. Methods like \cite{rebuffi2017icarl, cha2021co2l, fini2022self} utilize predictions from prior models as pseudo-labels for training on current tasks. 

\textbf{Contrastive Pretraining} In the realm of self-supervised learning (SSL), contrastive learning has emerged as a key technique. Unimodal methods like \cite{chen2020simple, yuan2022provable} create positive pairs from augmented input data, while bimodal methods such as CLIP \cite{radford2021learning, yuan2022provable} treat different modalities (e.g., image and caption) as positive pairs. Unimodal methods like \cite{cha2021co2l} adapt pretrained models using memory banks, while \cite{fini2022self} uses SSL objectives for cross-task knowledge transfer. Bimodal methods, like CLIP, have shown strong performance in both zero-shot and fine-tuning settings \cite{thengane2022clip, goyal2023finetune}, and recent studies explore their potential in continual learning contexts \cite{thengane2022clip, garg2024ticclip, cha2021co2l, cui2023generalized, li2024contrastive}.

\section{Methods}

\textbf{Notation.} Let $E_1, E_2$ denote the image encoder and text encoder respectively, parameterized by $\w$. A datasets $\D$ consists of $\T$ tasks where each task contains a subset of the dataset $\D^t$ where $\D^t \cap D^{t'} = \emptyset, \forall t' \neq t$ and $N_t = |\D^t|$.

\textbf{Class Incremental Learning} In class incremental learning, new tasks come with new classes.  The ultimate goal is to continually build a classification model for all classes. In other words, the model should not only acquire the knowledge from the current task $\D^t$ but also preserve the knowledge from former tasks. After each task, the trained model is evaluated over $\D^t_{test1} = \{(\x_i,y_i)\}$, $y_i \in \Y_t = Y_1 \cup ... Y_t$ and all the previously measured task $\D^b_{test} = \{(\x_i,y_i)\}$, for $b=1, ..., t-1$

\textbf{Domain Incremental Learning} In domain incremental learning, the goal is to update a model given some new data from another domain with the same set of labels. After being trained on tasks $t$, the model is evaluated on $\D^b_{test} = \{(\x_i,y_i)\}$, $y_i \in Y_t$.

\subsection{Bimodal Contrastive Continual Learning}

CLIP models, as shown in recent studies  \cite{radford2021learning, yuan2022provable}, possess the ability to process both image and text inputs by learning a joint embedding across modalities. Their impressive performance on image tasks without task-specific training is largely due to the contrastive learning objectives used during training. Moreover, encoding labels with the text encoder further boosts classification performance \cite{goyal2023finetune}.

Building on CLIP models' ability to jointly encode labels and images improves resistance to catastrophic forgetting and enhances adaptability to new data. To extend CLIP for CL, we propose a bimodal contrastive learning objective tailored to the class-incremental setting. The contrastive objective during each task is defined as:

\begin{equation}
 \begin{aligned}
\L_{contrastive} = &-\frac{1}{N_t + |\M_t|}\sum_{\x_i \in \D^t \cup \M_t} \log \frac{\exp((E_1(\w_t,\x_i)^T E_2(\w_t, y_i))/\tau)}{\sum_{y_j\in\D^t\cup \M_t} \exp((E_1(\w_t,\x_i)^T E_2(\w_t, y_j)) / \tau)} \\
&-\frac{1}{N_t + |\M_t|}\sum_{y_i \in \D^t \cup \M_t}\log \frac{\exp((E_2(\w_t,y_i)^T E_1(\w_t, \x_i))/\tau)}{\sum_{\x_j \in\D^t \cup \M_t} \exp((E_2(\w_t,y_i)^T E_1(\w_t, \x_j))/\tau)},
\end{aligned}   
\end{equation}
Here, $\D_t$ represents data for the current task, and $\M_t$ is a memory bank storing past task samples. Labels $y_i$ are encoded as text using $E_2$. To address computational constraints, we maintain a constant memory size, keeping an equal number of randomly sampled examples per class.

A key challenge in optimizing this objective arises from the summation over the entire dataset for contrastive terms:
\begin{align}
    g_I(\w, \x_i, \D^t\cup\M^t) &= \sum_{y_j\in\D^t\cup \M_t} \exp((E_1(\w_t,\x_i)^T E_2(\w_t, y_j)) / \tau)  \\
    g_T(\w, y_i, \D^t\cup\M^t) &= \sum_{\x_j \in\D^t \cup \M_t} \exp((E_2(\w_t,y_i)^T E_1(\w_t, \x_j))/\tau) 
\end{align}
      
To reduce the computational cost, we use moving average estimators $u^I_i$ and $u^T_i$ for $g_I$ and $g_T$. The gradient estimator is then computed using a mini-batch $\B$ as:

\begin{equation}
    \mathbf{m} = - \frac{1}{|\B|} \sum_{\x_i \in\B} \nabla (E_1(\w_t,\x_i)^T E_2(\w_t, y_i)) \\
    +\frac{\tau}{2|\B| u^I_i} \nabla g_I(\w, \x_i, \B) + \frac{\tau}{2|\B| u^T_i}\nabla g_T(\w, y_i, \B)
    \label{eq:gcl_grad_est}
\end{equation}

This method, which maintains a moving average across tasks, allows information from prior tasks to carry forward, enhancing CL. We call this approach the Global Contrastive Loss (GCL).

\subsection{Group Distributionally Robust Optimization}
\label{methods}

Due to the fixed memory size, after completing each task, we reduce the number of examples per class to accommodate new ones. This leads to an imbalance between previous and current task data distributions. While our Global Contrastive Loss (GCL) is effective for standard classification tasks, it doesn't handle these imbalances well. To address this, we introduce a group distributionally robust objective (DRO) that assigns greater weight to classes with higher losses during training.

We first define a contrastive loss for a specific class k as:
\begin{equation}
    h_k = \frac{1}{2 n_k}\sum_{i=1}^{n_k} (\tau \log g_1(\w, \x_i, \D^t\cup\M^t )+ \tau \log g_2(\w, y_i, \D^t\cup\M^t))
\end{equation}

where $g_1$ and $g_2$ are computed for negative samples and are influenced by a pairwise squared hinge loss. This formulation improves learning, especially in the context of partial AUC loss \cite{DBLP:conf/icml/ZhuLWWY22}.

The group DRO objective is then $\min_{\w}\max_{\p \in \Delta}\sum_{i=0}^{K_t} p_k h_k - \lambda \text{KL}(\p, 1/K_t)$ or equivalently,

 \begin{align}
     \min_\w \lambda\log\frac{1}{K_t}\sum_{k=1}^{K_t} \exp\left(\frac{h_k}{\lambda}\right)
     \label{eq:loss}
 \end{align}

This objective increases the weight for harder classes (those with higher losses) to reduce the imbalance.

In its compositional form, the DRO involves nested functions, making gradient estimation challenging. To address this, we apply a method from Stochastic Compositional Optimization, maintaining moving average estimators for the loss terms and contrastive components. These estimators allow us to efficiently compute the gradient using mini-batches:
 
 \[
 \frac{1}{v|\B_c|}\sum_{c_k\in \B_c}\exp \left(\frac{u_{c_k}}{\lambda} \right)\frac{1}{2|\B_k|}\sum_{(\x_i,y_i) \in \B_k}\left(\tau \frac{1}{u^I_i}\nabla g_1 + \tau \frac{1}{u^T_i} \nabla g_2 \right)
 \]
This approach ensures robust handling of imbalanced data distributions while efficiently optimizing the DRO objective.

 \begin{algorithm}[t]{\hspace*{-0.5in}}
    \centering
    \caption{The GDRO Method for Continual Learning of CLIP models}
    \label{alg:dro-kl}
    \begin{algorithmic}[1]  
    \STATE Set $\u^0=0, v^0 = 0$ and initialize $\w$
    \FOR {$t = 1,\ldots, T$}
    \STATE Sample a batch $\B$ 
    \STATE For each class $c_k \in \B$, sample a minibatch of data points denoted by $\B_k$.
    \STATE For each $c_k \in \B_c$, update $u^{\mathbf{I} (j)}_k =(1-\gamma)u^{\mathbf{I}(j-1)}_{k} + \gamma g_1(\w, \x_i, \D^t\cup\M^t) $
    \STATE For each $c_k \in \B_c$, update $u^{\mathbf{T} (j)}_k =(1-\gamma)u^{\mathbf{T}(j-1)}_{k} + \gamma g_2(\w, y_i, \D^t\cup\M^t) $
    \STATE For each $c_k \in \B_c$, update $u_{c_k}^{(j)} =(1-\gamma)u_{c_k}^{(j-1)} + \gamma h_k $
    \STATE Let $v^{(j)} = (1-\gamma)v^{(j-1)} + \gamma\frac{1}{K}  \sum_{i=1}^K \exp \left(\frac{u_k^{t}}{\lambda} \right)$
    \STATE Compute a gradient estimator $\nabla_j$ by 
          \[
            \frac{1}{v|\B_c|}\sum_{c_k\in \B_c}\exp \left(\frac{u_{c_k}}{\lambda} \right)\frac{1}{2|\B_k|}\sum_{(\x_i,y_i) \in \B_k}\left(\tau \frac{1}{u^I_i}\nabla g_1 + \tau \frac{1}{u^T_i} \nabla g_2 \right)
         \]
     \STATE Update $\v_{j}=(1-\beta_1)\v_{j-1} + \beta_1 \nabla_j$
     \STATE Update $\w_{j+1}   =\w_j - \eta_1  \v_j$ (or Adam-style)
    \ENDFOR
    \end{algorithmic}
\end{algorithm}

\section{Experiments}

In this section, we evaluate the effectiveness of the two methods referred to as GCL and GDRO in the class and domain incremental learning setting. For each of our experiments, we begin with a pretrained CLIP model \cite{cherti2023reproducible, king2023multimodal}. Our experiments are written in PyTorch \cite{paszke2017automatic} and are run on 4 NVIDIA RTX A5000 GPUs.

\subsection{Datasets}

We consider two class incremental datasets, namely CIFAR-100 and ImageNet. For domain incremental learning we evaluate our methods on DomainNet \cite{peng2019moment} which consists nearly 0.6 million images from 6 domains with 345 imbalanced classes. 

\subsection{Evaluation}

We measure the performance of a model by its ability to perform on the current task and the previous tasks that it has been previously trained on. In this section, we describe the metrics used throughout our experiments to evaluate our method. To show the learning process, after each stage, the trained model is evaluated over all classes that have already been trained, i.e., the t-th test set $\D^t_{test} = \{(\x_i,y_i)\}$, $y_i \in \Y_t = Y_1 \cup ... Y_t$. Denoted by $A_t$ the accuracy evaluated on $\D^t_{test}$ after stage $t$. 

\subsection{Baselines}

In our image experiments, we evaluate various baseline methods using a pretrained CLIP model with a VIT-B/16 vision encoder, starting with a zero-shot performance assessment to gauge prior knowledge. Our goal is to outperform this baseline with methods compared against benchmarks such as EWC \cite{kirkpatrick2017overcoming}, DER \cite{zhou2022model}, iCaRL \cite{rebuffi2017icarl}, Co2L \cite{cha2021co2l}, and FOSTER \cite{wang2022foster}, all utilizing the same pretrained encoder. Some dynamic expansion methods were omitted due to computational constraints.

For domain-incremental experiments on image datasets, we compare our approach with CaSSLe \cite{fini2022self}, focusing on the supervised contrastive objective for optimal results.

We ensure fair comparison by using consistent weight decay, batch size, and optimizer settings across methods, while fine-tuning the learning rate and number of epochs. For our DRO method, we additionally tune hyperparameters such as $\gamma$, $\lambda$, and the margin. We also vary memory sizes to test the effectiveness of our methods under different conditions.

\subsection{Class Incremental Learning}

\begin{figure}[ht]
    \centering
    \includegraphics[width=\linewidth]{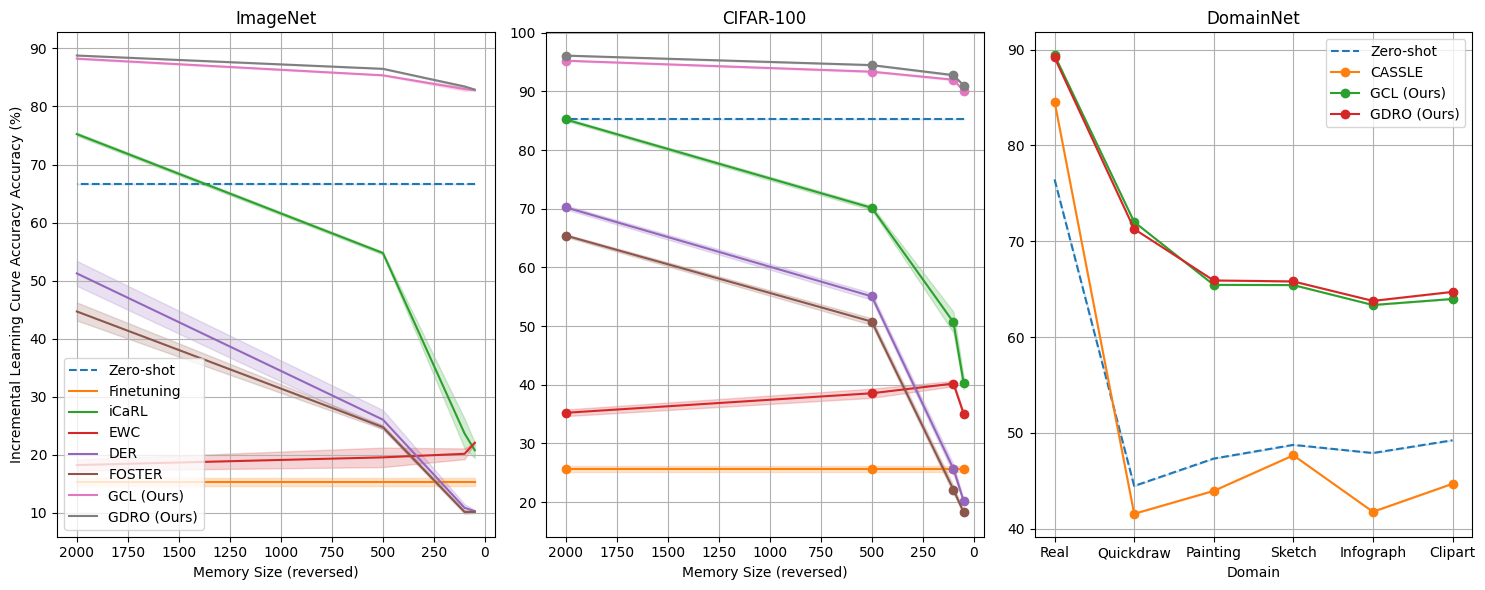}
    \caption{We report the mean and standard deviation of incremental learning curve accuracy over 3 runs on ImageNet1k at different memory sizes.}
    \label{fig:results}
    \vspace{-10px}
\end{figure}

\textbf{ImageNet1k Data} We further test our approach on the ImageNet1K dataset, splitting it into 10 tasks with 100 classes each. Due to the larger number of classes, we evaluate the methods with larger memory sizes. A finetuning baseline is also included, where the model is trained on all available data to establish an upper performance bound. Results are illustrated in Figure \ref{fig:results}.

Our methods significantly outperform others across all memory sizes. Notably, unimodal contrastive approaches like Co2L \cite{cha2021co2l} experience a sharp performance drop as memory size decreases. This is because Co2L relies on a self-supervised contrastive objective and requires labeled data from the memory bank for downstream tasks, which is limited when memory size is small.

\textbf{CIFAR-100 Data} We evaluate our method in a class-incremental learning (CIL) setting on the CIFAR-100 dataset, which is split into 10 tasks of 10 classes each. We assess performance across various memory sizes and report the accuracy after the final task. In addition, we compare our GDRO method with a baseline where finetuning is done solely with cross-entropy loss at each new task. Results are shown in Figure \ref{fig:results}.

As memory size decreases, our method performs comparably to zero-shot evaluation, indicating that while some forgetting occurs, our method maintains solid performance as it progresses through tasks. When comparing the contrastive method with the DRO method, we observe that the contrastive method performs better with larger memory sizes, but its performance drops significantly when no memory is available. In contrast, the DRO method maintains more stable performance under memory constraints.

\subsection{Domain Incremental Learning}

We evaluate our methods in the domain incremental learning (DIL) setting, beginning with the image-based DomainNet dataset. Accuracy is assessed after each task as performance on all prior tasks, and we also report the model's zero-shot performance before any training. Results are shown in Figure \ref{fig:results}.

Both of our methods outperform the baseline zero-shot results. As seen in our CIL experiments, contrastive CL methods like CaSSLe \cite{fini2022self} struggle to retain knowledge from previous tasks due to the absence of a memory bank, as it relies on a self-supervised objective at each step. In contrast, our DRO objective outperforms the GCL method after completing all tasks, demonstrating better retention and adaptability.

\section{Conclusion and Discussion}

We propose two methods using bimodal contrastive learning to jointly embed labels and input data for CL. The first incorporates label embeddings with a memory buffer to retain past task knowledge, while the second dynamically reweights harder examples to address class imbalance in the buffer.

Using a pretrained CLIP vision encoder, we evaluate these methods in class-incremental and domain-incremental learning on image datasets. Our contrastive method excels with a larger memory buffer, while dynamic reweighting proves most effective with a smaller buffer.

The results show that embedding both input data and labels reduces forgetting more effectively than linear classifiers. Reweighting classes enhances retention, especially with limited memory, highlighting the benefits of multimodal learning and adaptive weighting for CL in dynamic environments.



\bibliographystyle{plain}
\bibliography{Styles/references}

\end{document}